\documentclass[runningheads]{llncs}
\usepackage[T1]{fontenc}
\usepackage{graphicx,verbatim}
\usepackage{booktabs}
\usepackage{multirow}
\usepackage{amsmath}
\usepackage{amssymb}
\usepackage{caption}
\usepackage{gensymb}
\usepackage{marvosym}
\usepackage{ifsym}
\usepackage[colorlinks=true, citecolor=blue, linkcolor=blue]{hyperref}

\newcommand{\tabref}[1]{Table \ref{#1}}
\newcommand{\figref}[1]{Fig. \ref{#1}}
\newcommand{\secref}[1]{Sec. \ref{#1}}

\newcommand{\eg}{{\emph{e.g.}}}

\makeatletter
\newcommand{\printfontsize}{\f@size pt}
\makeatother

\def\ourmodel{Land-Reg}
\begin{document}
\title{Preoperative-to-intraoperative Liver Registration for Laparoscopic Surgery via Latent-Grounded Correspondence Constraints}

\titlerunning{Land-Reg for Deformable Liver Registration}
%

\author{Ruize Cui\inst{1} \and
Jialun Pei\inst{2(\textrm{\Letter})} \and
Haiqiao Wang\inst{2} \and
Jun Zhou\inst{1}  \and
Jeremy Yuen-Chun Teoh\inst{2} \and
Pheng-Ann Heng\inst{2}\and 
Jing Qin\inst{1}}

\authorrunning{R. Cui et al.}

\institute{The Hong Kong Polytechnic University, Hong Kong, China \and
The Chinese University of Hong Kong, Hong Kong, China  \\
\email{peijialun@gmail.com}}
  
\maketitle              
\begin{abstract}
In laparoscopic liver surgery, augmented reality technology enhances intraoperative anatomical guidance by overlaying 3D liver models from preoperative CT/MRI onto laparoscopic 2D views. However, existing registration methods lack explicit modeling of reliable 2D-3D geometric correspondences supported by latent evidence, leading to limited interpretability and potentially unstable alignment in clinical scenarios. 
In this work, we introduce~\ourmodel, a correspondence-driven deformable registration framework that explicitly learns latent-grounded 2D–3D landmark correspondences as an interpretable intermediate representation to bridge cross-modal alignment. For rigid registration,~\ourmodel~embraces a Cross-modal Latent Alignment module to map multi-modal features into a unified latent space. Further, an Uncertainty-enhanced Overlap Landmark Detector with similarity matching is proposed to robustly estimate explicit 2D-3D landmark correspondences. For non-rigid registration, we design a novel shape-constrained supervision strategy that anchors shape deformation to matched landmarks through reprojection consistency and incorporates local-isometric regularization to alleviate inherent 2D-3D depth ambiguity, while a rendered-mask alignment enforces global shape consistency.
Experimental results on the P2ILF dataset demonstrate the superiority of our method on both rigid pose estimation and non-rigid deformation. Our code will be available at \url{https://github.com/cuiruize/Land-Reg}.


\end{abstract}
\section{Introduction}

\begin{figure}[t!]
\centering
\includegraphics[width=0.96\textwidth]{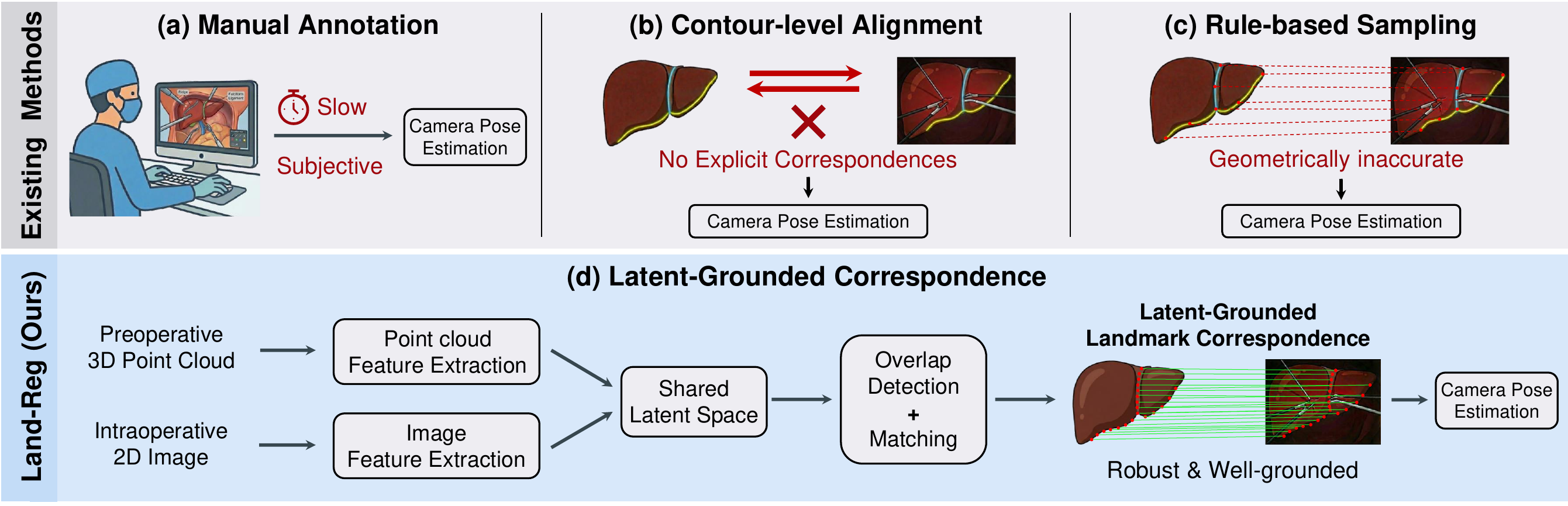}
\caption{Comparison of landmark-based rigid liver registration methods.}
\label{fig1}
\end{figure}

Laparoscopic liver surgery constitutes a prevalent treatment for hepatic malignancies, yet the restricted field-of-view and continuously changing viewpoints significantly hinder accurate intraoperative navigation~\cite{fretland2018laparoscopic,giannone2024robotic}. 
Augmented reality (AR) guidance enhances operating precision by overlaying patient-specific 3D models reconstructed from preoperative CT/MRI onto intraoperative laparoscopic videos, integrating preoperative anatomical context directly into the surgical field in a spatially coherent manner~\cite{ramalhinho2023value}. 
A fundamental prerequisite for reliable AR guidance involves accurate 2D–3D registration between the preoperative model and each intraoperative frame. 
However, achieving robust registration under laparoscopic conditions remains challenging due to partial visibility, frequent occlusions, drastic viewpoint variations, and substantial non-rigid liver deformation induced by respiration and surgical manipulations~\cite{espinel2021using,lei2024epicardium,padovan2022deep}.

Existing 2D-3D liver registration methods predominantly utilize landmarks as effective biomarkers to bridge 2D-3D modalities~\cite{plantefeve2014automatic}.
Early rigid registration methods~\cite{ozgur2018preoperative,robu2018global} rely on manual landmark annotation or pose initialization (\figref{fig1} (a)), limiting automation and introducing user variability. Recent learning-based methods~\cite{p2ilf,gadoux2025automatic,koo2022automatic,lmr,mhiri2025neural,zhang2025deep} automatically detect anatomical structures and estimate rigid camera pose by minimizing landmark reprojection errors (\figref{fig1} (b)). 
Despite improved automation, these approaches treat registration as global parameter regression or contour-level alignment, without explicitly establishing 2D–3D geometric correspondences. As a result, the alignment lacks interpretable geometric constraints and becomes unstable in clinical scenarios. 
Alternative strategies~\cite{p2ilf,opt} enforce point-level matching via pre-defined rule-based correspondence sampling along detected landmark curves (\figref{fig1} (c)). These heuristic strategies assume consistent curve parameterization (\eg, uniform sampling) and neglect true anatomical overlap. 
In this regard, we formulate latent-grounded landmark correspondences (\figref{fig1} (d)) to impose rigorous geometric constraints for interpretable and stable rigid registration. 

Beyond rigid alignment, precise AR guidance requires adaptation to intraoperative non-rigid deformation. Estimating 3D deformation from a monocular view is fundamentally ill-posed: multiple distinct 3D deformations can generate nearly identical 2D projections~\cite{laurentini2002visual,salzmann2010linear}, and purely contour- or landmark-driven supervisions~\cite{adagolodjo2017silhouette,koo2017deformable,plantefeve2014automatic} may produce depth-ambiguous solutions featuring unrealistic local distortions. 
To alleviate the inherent 2D-3D depth ambiguity, the deformation strategy requires integrating local-shape constraints in supervision.

This work proposes~\ourmodel, a two-stage correspondence-driven paradigm for preoperative-to-intraoperative liver registration.
The insight involves explicitly formulating latent-grounded correspondences (\figref{fig1} (d)) to ensure interpretable and reliable 2D–3D alignment. 
For rigid registration, we introduce Cross-modal Latent Alignment (CLA) to map multi-modal features into a unified latent space, alongside an Uncertainty-enhanced Overlap Landmark Detector (UOLD) to identify overlapping landmarks. 
By performing similarity matching, \ourmodel~establishes anatomically grounded and interpretable 2D-3D landmark correspondences, providing geometric constraints for camera pose estimation via EPnP~\cite{epnp} with RANSAC~\cite{ransac}. 
During the deformation, a shape-constrained supervision strategy is designed to alleviate 2D-3D depth ambiguity, containing a correspondence reprojection error $\mathcal{L}_{cre}$ to anchor matched landmarks and a local-isometric regularizer $\mathcal{L}_{iso}$ to suppress local distortions, as well as a rendered-mask alignment $\mathcal{L}_s$ for global shape calibration. 
Extensive experiments on the P2ILF dataset demonstrate that~\ourmodel~significantly improves both rigid pose estimation and non-rigid deformation reconstruction.

\section{Methodology}

\begin{figure}[t!]
\includegraphics[width=0.98\textwidth]{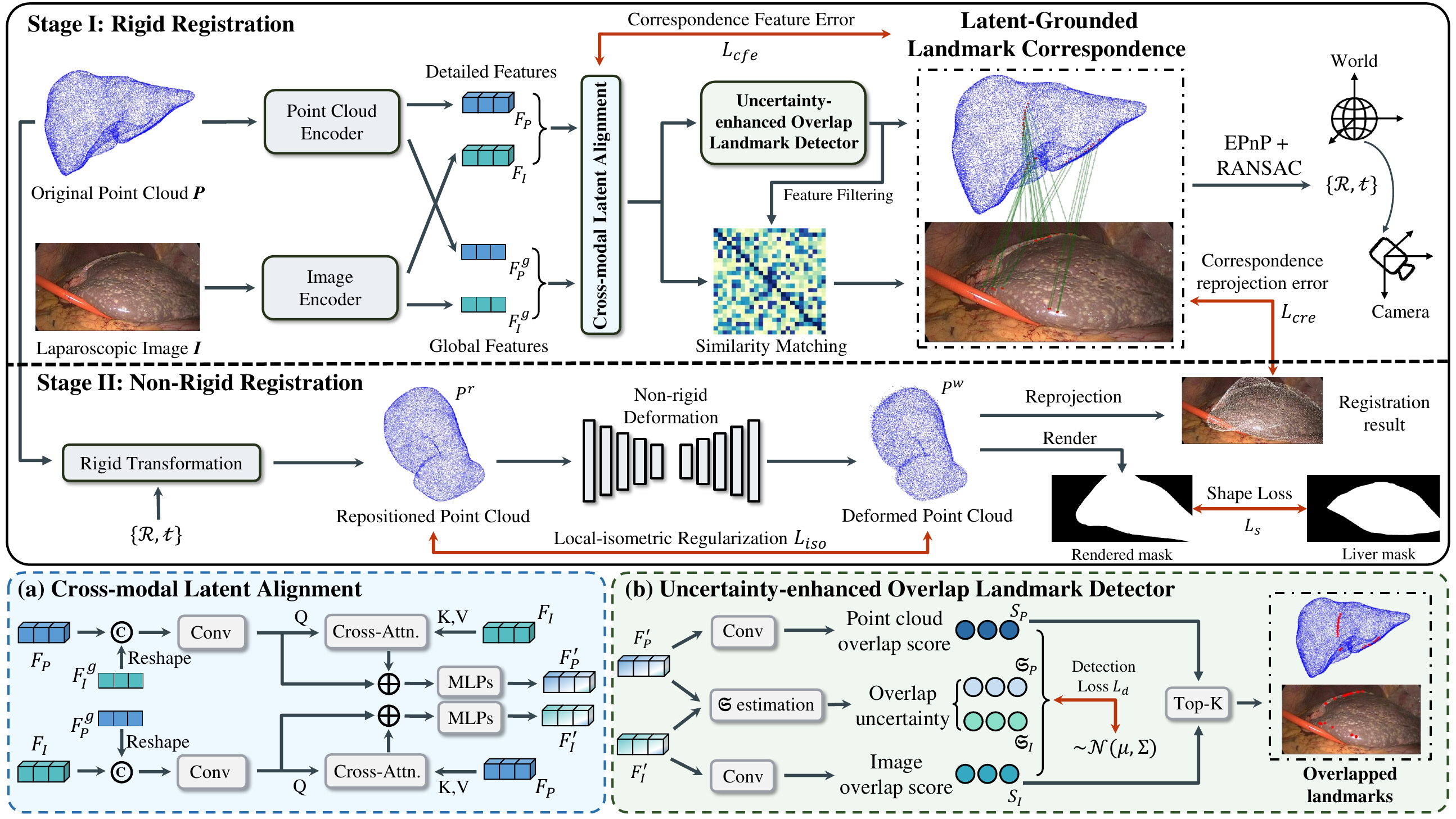}
\caption{Architecture of~\ourmodel~for 2D-3D liver registration.}
\label{overview}
\end{figure}

\subsection{Overview}
Given an intraoperative laparoscopic frame $\textbf{I}\in \mathbb{R}^{W \times H \times 3}$ and the corresponding preoperative liver point cloud $\textbf{P}\in \mathbb{R}^{N\times3}$, preoperative-to-intraoperative liver registration aims to identify (\romannumeral1) a rigid camera pose $\textbf{T}=[\textbf{R},\textbf{t}]$, where $\textbf{R}\in{SO}(3)$ and $\textbf{t}\in \mathbb{R}^{3}$, and (\romannumeral2) a non-rigid deformation field $\mathbf{\Phi} \in \mathbb{R}^{N\times3}$ that wraps $\textbf{P}$ to the intraoperative shape. The wrapped model $\textbf{P}^w$ is then projected onto \textbf{I} using the camera intrinsic matrix $\textbf{K}\in \mathbb{R}^{3\times3}$ to obtain the registration result. 

As illustrated in~\figref{overview}, \ourmodel~follows a two-stage design: Stage \uppercase\expandafter{\romannumeral1} for rigid registration and Stage \uppercase\expandafter{\romannumeral2} for non-rigid registration. 
In Stage \uppercase\expandafter{\romannumeral1}, we extract point-wise and pixel-wise features from $\textbf{P}$ and $\textbf{I}$ via dedicated backbones, and map them into a shared latent space through Cross-modal Latent Alignment (CLA) module. Subsequently, the Uncertainty-enhanced Overlap Landmark Detector (UOLD) identifies reliable overlap landmark points in both modalities. 
Following feature filtering at these candidates, cosine similarity matching within the unified latent space is conducted for 2D-3D landmark correspondences. The camera pose $\textbf{T}$ is estimated by utilizing EPnP~\cite{epnp} with RANSAC~\cite{ransac} for outlier rejection. 
In Stage \uppercase\expandafter{\romannumeral2}, we transform $\textbf{P}$ into the camera coordinate system using $\textbf{T}$ and estimate non-rigid liver deformation to obtain $\textbf{P}^w$. We also propose a novel shape-constrained supervision strategy to suppress 2D-3D depth-ambiguity.

\subsection{Rigid Registration via Latent-Grounded Correspondence}
To achieve reliable and interpretable rigid registration, \ourmodel~models 2D-3D landmark correspondences grounded in the unified latent space as geometric constraints to estimate the camera pose.

\noindent\textbf{Cross-modal Latent Alignment.}
Given $\textbf{P}$ and $\textbf{I}$, we utilize PointNet++~\cite{pointnet++} and ResNet~\cite{resnet} to extract detailed point cloud feature $F_P\in \mathbb{R}^{N \times d}$ and image feature $F_I\in \mathbb{R}^{\frac{W}{4}\times\frac{H}{4}\times d}$, where $d$ denotes the feature dimension, and apply pooling operations on $F_P$ and $F_I$ to obtain global features $F_P^g\in \mathbb{R}^d$ and $F_I^g\in \mathbb{R}^d$. 
CLA map $F_P$ and $F_I$ into a shared latent space, supporting overlap landmark detection and correspondence modeling. As shown in~\figref{overview} (a), CLA contains two structurally symmetric branches for point cloud and image features. By interacting with cross-modal global features by convolutions and cross attention, our model can generate latent-unified features $F_P^{'} \in \mathbb{R}^{N \times d}$ and $F_I^{'} \in \mathbb{R}^{\frac{W}{4}\times\frac{H}{4}\times d}$.

\noindent\textbf{Uncertainty-enhanced Overlap Landmark Detector.} 
Since intraoperative liver surface may partially overlap with its preoperative 3D model, UOLD is proposed to detect overlapping liver landmarks.
Given $F_P^{'}$ and $F_I^{'}$, UOLD predicts point-wise and pixel-wise overlap likelihoods $S_P \in \mathbb{R}^{N}$ and $S_I \in \mathbb{R}^{\frac{W}{4}\times\frac{H}{4}}$ using lightweight detection heads, indicating how likely each 3D point or 2D pixel belong to the overlapping region.
To counter unreliable scores caused by outliers and landmark distribution imbalance, it is critical to identify anatomically individuals that are supposed to attract more attention for promoting detection accuracy and rejecting artifacts~\cite{kendall2017uncertainties}. 
In this regard, we employ two prediction heads to estimate point-wise and pixel-wise uncertainty terms $\mathfrak{S}_P \in \mathbb{R}^{N}, \mathfrak{S}_I \in \mathbb{R}^{\frac{W}{4}\times\frac{H}{4}}$, and use them to optimize $S_P$, $S_I$ with Gaussian negative log-likelihood $\mathcal{L}_{nll}$. Assume $O_P, O_I$ are the overlapping landmarks based on ground truth, we sample equivalent non-overlapping points and pixels $U_I, U_P$ to construct $\mathcal{Q}\in \{O_I,  O_P, U_I, U_P\}$, the detection Loss ${L}_d$ is defined as
\begin{equation}
\begin{gathered}
    \mathcal{L}_{nll}^\vartheta = \frac{1}{n}\sum_{i=1}^n\Big(\frac{(Y_\vartheta^i-S_\vartheta^i)^{2}}{2\mathfrak{S}_\vartheta^i} + \frac{1}{2}\log(\mathfrak{S}_\vartheta^i)\Big), \quad\mathcal{L}_{d} =\frac{1}{4}\sum_{\vartheta \in \mathcal{Q}}\mathcal{L}_{nll}^\vartheta,
\end{gathered}
\end{equation}
where $n$ is the number of samples in $\vartheta$, and $Y\in\{0,1\}$ is ground truth. $\mathcal{L}_d$ encourages precise overlap scores while suppressing ambiguous samples.

\noindent\textbf{Landmark Correspondence Modeling and Rigid Registration.} 
After obtaining overlapping landmarks, we filter representations of selected candidates and establish 2D-3D correspondences by mutual nearest neighbor matching based on cosine similarity matching. 
We apply EPnP with RANSAC to estimate the camera pose $\textbf{T}$ and reject the outliers simultaneously. Meanwhile, a contrastive learning strategy is designed to enhance the discriminativeness of paired correspondence features in the unified latent space:
\begin{equation}
    \mathcal{L}_{cfe} = \frac{1}{m}\sum_{i=1}^m (2-\mathfrak{s}_o^i + \mathfrak{s}_u^i),
\end{equation}
where $\mathfrak{s}_o^i$ denotes the similarity between features of the $i^{th}$ pair of matched landmark points, while $\mathfrak{s}_u^i$ denotes the largest similarity value between features of the $i^{th}$ 3D overlapping landmark and unmatched 2D overlapping landmark pixels. $m$ represents the number of landmark correspondences. Our final loss function for rigid registration is
\begin{equation}
    \mathcal{L}_{rigid} = \mathcal{L}_d + \lambda\mathcal{L}_{cfe},
\end{equation}
where $\lambda$ is the balancing parameter and empirically set to 0.5 in our framework.

\subsection{Shape Constrained Deformable Registration}
To alleviate the negative effects caused by depth-ambiguity, Stage \uppercase\expandafter{\romannumeral2} combines global shape alignment with correspondence reprojection consistency and local isometry priors for supervision, suppressing depth-ambiguous deformation. Given the predicted camera pose $\textbf{T}$ in Stage \uppercase\expandafter{\romannumeral1}, we transform $\textbf{P}$ from the world coordinate system into the camera coordinate system to obtain the repositioned point cloud~$\textbf{P}^r = \textbf{RP} + \textbf{t}$. Then a deformable autoencoder~\cite{autoencoder} is adapted to estimate deformation $\mathbf{\Phi}$ and obtain the wrapped 3D model $\textbf{P}^w = \textbf{P}^r + \mathbf{\Phi}$. 
To prevent model collapse, an initial warm-up phase is applied to train the autoencoder to reconstruct $\textbf{P}^r$ with MSE Loss. Following, the MSE Loss is replaced with the proposed shape-constrained supervision strategy to supervise deformation estimation, comprising a Shape Loss $\mathcal{L}_s$, a Correspondence Reprojection Error $\mathcal{L}_{cre}$, and a Local-isometric Regularization $\mathcal{L}_{iso}$. 
Inspired by~\cite{landmarkfree}, $\mathcal{L}_s$ utilizes a Dice loss to align the predicted mask rendered from $\textbf{P}_w$ with the segmented liver surface mask. This term maintains the global shape consistency between the deformed model and the intraoperative liver surface. However, because $\mathcal{L}_{s}$ only focuses on the alignment of 2D projections, it is inherently under-constrained in 3D and affected by 2D-3D depth ambiguity. Therefore, $\mathcal{L}_{cre}$ and $\mathcal{L}_{iso}$ are incorporated to strengthen the supervision to overcome depth ambiguity. Since we have predicted the landmark correspondences in Stage \uppercase\expandafter{\romannumeral1}, we minimize the landmark correspondence reprojection error $\mathcal{L}_{cre}$ to ensure the consistency of matched landmarks under deformation. Let $\{X_i, x_i\}_{i=1}^{M}$ be the set of $M$ matched landmark correspondences, we can formulate $\mathcal{L}_{cre}$ by,
\begin{equation}
    \mathcal{L}_{\text{cre}}=\frac{1}{M}\sum_{i=1}^{M}\left\|\pi(X_i; \textbf{R,t,K})-x_i\right\|_2^2,
\end{equation}
where $\pi$ denotes 3D-2D projection. In addition, $\mathcal{L}_{iso}$ is introduced to maintain locally rigid structure of the point cloud and prevent local stretching. Specifically, we apply the farthest point sampling to build a subset $\textbf{P}_{f}$ with $m$ points and construct a fixed k-nearest-neighbor graph $\mathcal{G}(i)$ on each point $i$ in $\textbf{P}_f$. During training, we penalize pairwise local distortion within $\mathcal{G}(i)$ with $\mathcal{L}_{iso}$:
\begin{equation}
    \mathcal{L}_{iso}=\frac{1}{m(m-1)}\sum_{i=1}^{m}\sum_{\substack{j \in \mathcal{G}(i) \\ j \neq i}}\Bigl(\lVert \hat{\mathbf{p}}_i-\hat{\mathbf{p}}_j \rVert_2^2-\lVert \mathbf{p}_i-\mathbf{p}_j \rVert_2^2 \Bigr)^2,
\end{equation}
where $\mathbf{p}_i,\mathbf{p}_j$ denote the points from $\textbf{P}_f$, while $\hat{\mathbf{p}}_i,\hat{\mathbf{p}}_j$ denote the corresponding deformed points. The final loss function for Stage \uppercase\expandafter{\romannumeral2} is:
\begin{equation}
    \mathcal{L}_{def} = \mathcal{L}_{s} + \lambda_{cre}\mathcal{L}_{cre} + \lambda_{iso}\mathcal{L}_{iso},
\end{equation}
where $\lambda_{cre}, \lambda_{iso}$ are balancing parameters. During inference, $\mathcal{L}_{def}$ is utilized within a run-time optimization (RTO) strategy~\cite{gadoux2025automatic} for deformation refinement.

\section{Experiments}


\begin{table}[t]
{\fontsize{7}{7.5}\selectfont  
    \renewcommand{\arraystretch}{1.2}
    \centering
    \captionof{table}{Quantitative comparisons of registration methods on the P2ILF dataset.}
    \resizebox{\linewidth}{!}{
        \begin{tabular}{l|ccc|ccc|c|c}
        \hline\hline
         \multirow{2}{*}{Methods} & \multicolumn{3}{c|}{Rigid Registration} & \multicolumn{3}{c|}{Non-rigid Registration} & \multirow{2}{*}{$\text{RRE} \downarrow$} & \multirow{2}{*}{$\text{RTE} \downarrow$} \\
         \cline{2-4} \cline{5-7}
         & $\text{Dice} \uparrow$ & $\text{TRE}_a \downarrow$ & $\text{TRE}_s \downarrow$ & $\text{Dice} \uparrow$ & $\text{TRE}_a \downarrow$ & $\text{TRE}_s \downarrow$ & & \\
        \hline
        BHL~\cite{p2ilf} & 55.58 & 203.88 & 101.50 & - & - & - & 94.36 & 493.48 \\
        Grasp~\cite{p2ilf} & \underline{66.64} & 348.39 & 293.22 & - & - & - & 99.29 & 134.01 \\
        NCT~\cite{p2ilf} & 56.55 & {117.98} & {99.17} & - & - & - & 78.56 & 90.83 \\
        UCL~\cite{p2ilf} & 32.21 & 174.05 & 147.87 & - & - & - & 247.86 & 191.36 \\
        VOR~\cite{p2ilf} & 28.59 & 443.89 & 375.09 & - & - & - & 225.79 & 202.42 \\
        \hline
        LMR~\cite{lmr} & 51.45 & 278.44 & 245.57 & 53.41 & 197.86 & 156.79 & 77.62 & 126.85\\
        Opt~\cite{opt} & 57.57 & 219.24 & 186.15 & 59.29 & 168.56 & 142.33 & 72.87 & 112.59 \\
        ADeLiR~\cite{gadoux2025automatic} & 62.27 & \underline{107.53} & \underline{92.58} & \underline{65.16} & \underline{89.14} & \underline{75.60} & \underline{62.49} & \underline{67.42} \\
        \hline
        \textbf{\ourmodel~(Ours)} & \textbf{69.21} & \textbf{90.51} & \textbf{80.28} & \textbf{75.38} & \textbf{76.46} & \textbf{63.67} & \textbf{45.52} & \textbf{42.26} \\
        \hline\hline
        \end{tabular}
        \label{results}
    }}
\end{table}

\begin{figure}[t!]
\centering
\includegraphics[width=\textwidth]{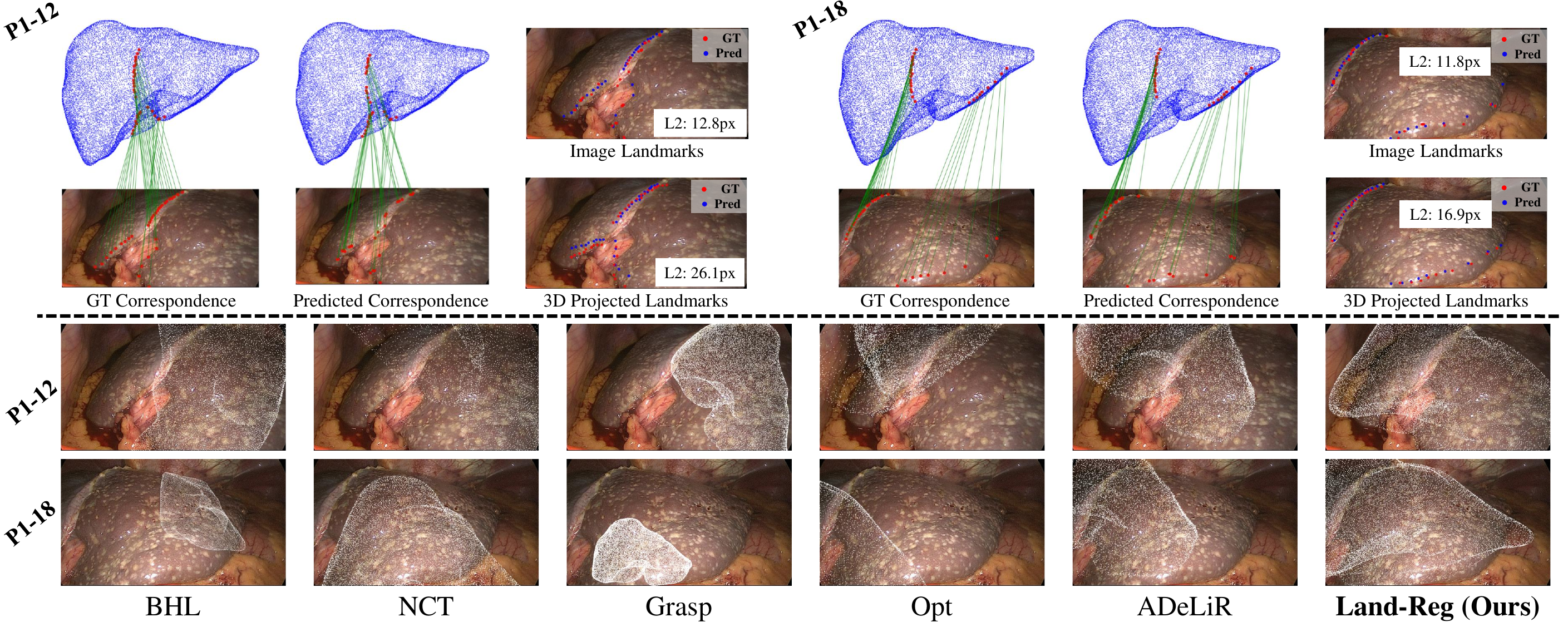}
\caption{Qualitative results from two randomly selected samples. Top: correspondence learning performance of~\ourmodel, including comparisons of correspondence and overlapping landmark detection between GT and our predictions. Bottom: registration results comparison of~\ourmodel~and competitive methods.} \label{vis}
\end{figure}

\subsection{Dataset and Pseudo Label Generation} \label{dataset}
We evaluate our method on the P2ILF~\cite{p2ilf} dataset, which provides preoperative 3D liver models, intraoperative 2D laparoscopic images, and corresponding landmark annotations for registration. Due to the inaccessibility of the official test set, only the P2ILF training set is used for experiments, which contains 167 laparoscopic images from nine patients with patient-specific 3D liver models, and we randomly select one patient case for testing and the others for training.

To obtain direct 2D-3D correspondence supervision, we apply arc-length aligned resampling to generate pseudo labels. For each ridge or ligament curve, we resample the sparse 3D landmark points to match the number of corresponding 2D pixels, yielding dense 2D-3D correspondences. Each original 3D landmark point was then associated with its arc-length aligned 2D pixel to construct pseudo correspondence labels for supervision. In addition, we also apply SAM2~\cite{sam2} to segment intraoperative liver masks for deformation supervision.

\subsection{Implementation Details}
We train and test~\ourmodel~on two NVIDIA RTX 3090 GPUs. We train 200 epochs with a batch size of 4 for both stages, and we run 50 epochs for warm-up in Stage \uppercase\expandafter{\romannumeral2}.
We adapt the Adam optimizer with an initial learning rate of 1e-4. We empirically set $\lambda_{cre} = 0.5$ and $\lambda_{iso} = 0.5$ for balancing parameters. For evaluation metrics, we follow~\cite{gadoux2025automatic} to measure the registration accuracy using the target registration error (TRE) in pixels (px) between the ground-truth 2D landmarks and the projections of registered 3D landmarks and summarize it by the mean ($\text{TRE}_a$) and standard deviation ($\text{TRE}_s$). We additionally compute the Dice score between the rendered liver mask and the ground truth to assess global shape calibration and use relative rotation error (RRE) in degrees ($\degree$) and relative translation error (RTE) in millimeter (mm) to evaluate camera pose estimation.

\subsection{Experimental Comparisons}
We compare our~\ourmodel~with five challenge methods in P2ILF~\cite{p2ilf} for rigid registration and follow~\cite{gadoux2025automatic} to compare with Opt~\cite{opt}, LMR~\cite{lmr} and ADeLiR~\cite{gadoux2025automatic} for deformable registration. 
As shown in~\tabref{results}, \ourmodel~outperforms all the competitors across all evaluation metrics in both registration tasks. Relative to the competitive baseline ADeLiR, our method enhances rigid reprojection performance by increasing 6.94\% on Dice, while decreasing 17.02 pixels on TRE$_a$, and 12.30 pixels on TRE$_s$. 
Furthermore, evaluating registration accuracy reveals reductions of 14.97$\degree$ on RRE and 25.16mm on RTE, demonstrating the effectiveness of enforcing latent-grounded correspondence constraints within the framework. Notably, we find that Grasp method presents a performance divergence, achieving an outstanding global reprojection performance of 66.64\% on Dice but the worst TRE$_a$ and TRE$_s$ outcomes and inferior registration accuracy. This discrepancy arises because Grasp exclusively utilize the rendered-mask alignment to constrain camera pose estimation, thus affected by 2D-3D depth ambiguity and obtain an inaccurate camera pose that can produce similar rendered-mask with the ground truth. 
In terms of non-rigid registration, the results indicate that our method brings the largest reprojection improvements compared to the rigid registration results across all deformable registration stage, which underscores the importance of suppressing depth-ambiguity during deformation learning. 
\figref{vis} exhibits the qualitative results. The top panel demonstrates the geometrically precise 2D-3D correspondence prediction of~\ourmodel, while the bottom one showcases our superiority on 2D-3D laparoscopic liver registration.

\subsection{Ablation Study}
\begin{figure}[t!]
\centering
\includegraphics[width=\textwidth]{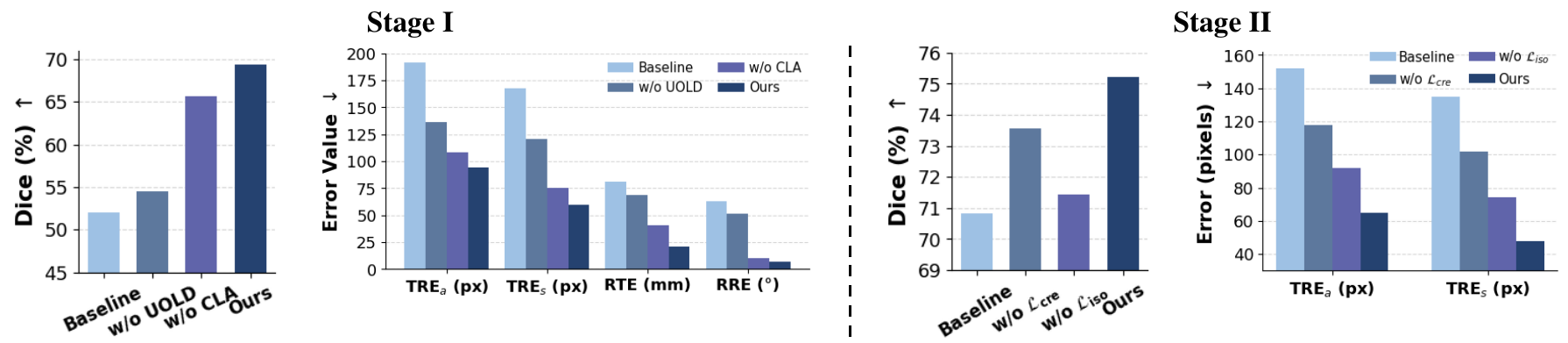}
\caption{Ablation results of crucial components proposed in~\ourmodel.} \label{abla}
\end{figure}

We conduct ablation studies on the crucial components of~\ourmodel~on the same test set outlined in~\secref{dataset}. In Stage \uppercase\expandafter{\romannumeral1}, the baseline replaces the CLA with a concatenation operation, removes UOLD, and incorporate all landmarks for similarity matching. As illustrated in the left of~\figref{abla}, each component contributes to our framework. Notably, integrating UOLD yields significant performance gains, demonstrating the importance of filtering overlapping landmarks to establish precise correspondences. In Stage \uppercase\expandafter{\romannumeral2}, the baseline is constructed by omitting $\mathcal{L}_{cre}$ and $\mathcal{L}_{iso}$ during training. The right of~\figref{abla} indicates that $\mathcal{L}_{iso}$ enhances global shape calibration by suppressing depth-ambiguous distortions, while $\mathcal{L}_{cre}$ effectively supports the reprojection consistency of matched landmarks. 


\section{Conclusion}
We present~\ourmodel, a two-stage correspondence-driven registration framework that explicitly models latent-grounded 2D-3D correspondences for interpretable liver alignment in laparoscopic surgery. We first introduce cross-modal latent alignment and uncertainty-enhanced overlap landmark detector to construct 2D-3D landmark correspondences grounded by latent representation-level evidence for robust camera pose estimation.
To suppress depth ambiguity in deformation estimation, we adopt a shape-constrained supervision strategy comprising landmark anchoring, local-isometric regularization and rendered-mask alignment.
Compared to cutting-edge methods,~\ourmodel~achieves more precise and interpretable registration, demonstrating its potential in surgical AR navigation. 

\bibliographystyle{splncs04}
\bibliography{ref}

\end{document}